\documentclass[a4paper,11pt]{article}

\usepackage{ucs}
\usepackage[utf8x]{inputenc}

\usepackage{times} 
\usepackage[T1]{fontenc}         
\usepackage{pslatex}
\usepackage[french,english]{babel}
\usepackage{hyperref}

\usepackage{graphicx}

\PrerenderUnicode{âĂ ̨e}

\title{Manuscripts in Time and Space: Experiments in 
	Scriptometrics on an Old French Corpus\thanks{%
		A digital appendix to this paper is available on Zenodo, \textsc{doi}: \href{https://doi.org/10.5281/zenodo.1117924}{10.5281/zenodo.1117924}. 
		My gratitude, for discussion on this subject over the years, goes to Frédéric Duval, Martin D. Gleßgen, Hans Goebl and Achim Stein.
		I also thank the anonymous reviewers for their insightful advice.
}}

\author{Jean-Baptiste Camps
\\[0.5cm] Centre Jean-Mabillon
\\École nationale des chartes | Paris Sciences \& Lettres
\\E-mail: \texttt{jbcamps@hotmail.com}}

\date{}

\begin{document}
\maketitle

\begin{abstract}
\noindent
Witnesses of medieval literary texts, preserved in manuscript, are layered objects, being almost exclusively copies of copies. This results in multiple and hard to distinguish linguistic strata -- the author's \textit{scripta} interacting with the \textit{scriptae} of the various scribes -- in a context where literary written language is already a dialectal hybrid. Moreover, no single linguistic phenomenon allows to distinguish between different \textit{scriptae}, and only the combination of multiple characteristics is likely to be significant \cite{goebl_franzosische_1995} -- but which ones? The most common approach is to search for these features in a set of previously selected texts, that are supposed to be representative of a given \textit{scripta}. This can induce a circularity, in which texts are used to select features that in turn characterise them as belonging to a linguistic area. 
To counter this issue, this paper offers an unsupervised and corpus-based approach, in which clustering methods are applied to an Old French corpus to identify main divisions and groups. Ultimately, scriptometric 
profiles are built for each of them.

\end{abstract} 

\thispagestyle{empty}

\section{Introduction}




Study on the diatopic variation of medieval French texts rests on the distinction proposed by 
Remacle \cite{remacle_probleme_1948} 
between \textit{scripta}, written language (
\textit{Schriftsprache}), and dialect, spoken language, the latter mostly inaccessible to us. 
Based on his study of 
Walloon
, this distinction was put forward as a mean to reconcile the difference he observed between the very characterized 
modern dialect and the medieval written texts from the area, presumably less marked by local traits. 
In the medieval \textit{scripta}, he argued, the distinctive traits inherited from spoken Walloon would be present only by mistake or ignorance. \enlargethispage{1\baselineskip} 
Consequently, he formulated the apparently self-contradictory hypothesis that ``1. the scripta was the result of a local development, 2. the scripta was a common language whose essential elements were found in most spoken dialects of the \textit{langue d'oïl}'' (my translation). 
%
This distinction 
is now commonly accepted  
though sometimes criticised because it sets in stone our inability to ever gain insights into the reality of medieval dialects 
\cite{dees_dialectes_1985}.
For the scholar who wants to date and localise the \textit{scripta} of medieval texts, this implies that he will face a language that was never spoken as such and the very building blocks of which might be made of elements taken from various dialectal areas, maybe even a \textit{koinè}, in which truly local traits are only marginal 
\cite[p.~40]{duval_francais_2009}
.

The exact reality of this notion of \textit{scripta} is still debated, but, as a working definition, we will take it as
the written language, practised by a restricted number of literates, around scriptural centres (e.g. chancelleries), and supposedly conceived to allow for a broader comprehension than oral dialects, but still containing traits that can be geographically assigned to a specific area. 
The possible connexion between the main modern dialectal areas (as delimited by modern dialectologists) and the geographical hold of medieval documentary \textit{scriptae} can be estimated due to the fact that administrative documents (charters, for instance) are usually dated (time and place date). It seems confirmed by Goebl's work \cite{goebl_lamenagement_2011}.

The case is even more complex for 
literary witnesses\footnote{%
	I define \textit{witness} as a given instance of a text, 
	as preserved in a particular document (usually, a manuscript) that is accessible to us. 
	See Duval \cite{duval_pour_2017} for an account on the meaning of the terms \textit{text} and \textit{witness} (``texte'' and ``témoin'') in (neo-lachamannian) textual criticism. 
	It allows me to distinguish between the more abstract work (e.g. the story of Roland and the battle at Roncevaux)
	and its expression in particular texts (i.e. the \textit{Chanson de Roland} or the \textit{Cân Rolant}), attested in witnesses (e.g. \textit{O}), preserved in documents (the ms. Digby~23).
}. 
While documentary texts (charters, wills,…) are practical documents, often of only local interest, 
most literary texts 
 were 
 made to be able to circulate through different linguistic areas, written by the more knowledgeable amongst the population, and influenced by the written codes of Latin \cite[p.~41]{duval_francais_2009}. 
Sociolinguistics 
played a part, as well as factors related to production of books, such as the implantation of workshops. 
Variation in prestige between dialects led to difference in behaviour among writers, 
up to the point where some \textit{scriptae} were judged distinctive of a genre, and its features imitated,  
like Western dialects or Picard for epic texts \cite
{bennett_normand_2003}.
Two scribes working in the same workshop but from different origin 
might produce a text with different features. As such, localising the \textit{scripta} of a witness does not mean as much finding its place of origin as identifying the linguistic inclinations of its writers \cite
{tyssens_typologie_1990}.
 But the major difficulty is of another nature yet: literary witnesses are layered objects, in which the language of the author interacts with 
each scribe's%
, up to the point where it is a very delicate task to assign any trait to a given layer, especially since any layer might already have included an alternation of forms or mixed forms \cite
{remacle_probleme_1948}.

As a consequence, it is very hard for dialectologists to determine isoglosses, or more precisely isographs \cite[p.~166]{monfrin_mode_2001}, that could clearly separate different \textit{scriptae}.
In fact, it is likely that no single trait can be used to define a \textit{scripta} \cite[p.~315]{goebl_franzosische_1995}
: 
most isographs are shared among several --~usually neighbouring~-- regions \cite[p.~65]{lusignan_langue_2004}. Even for the rare isographs that would be very distinctive, the information they provide is blurred by 
the hybrid nature of \textit{scriptae} or the stratification of textual witnesses. 
As a consequence, only a combination of traits, individually common with other \textit{scriptae}, in a given relative frequency, makes the distinction possible. This has led to an emphasis put on quantification,  
and eventually on statistical multivariate analysis \cite[p.~317]{goebl_franzosische_1995}. This approach is named ``dialectometry'' since Séguy \cite{seguy_dialectometrie_1973}, or, better in our case,  ``scriptometry''. It is defined by Goebl 
\cite[p.~60-61]{goebl_regards_2003}
as an alliance between linguistic geography and clustering, and it shares some similarities with, for instance, stylometry and other historical text analysis fields. More generally, it can be defined as \textit{the measure of scriptologic features}. 
As an exploratory approach, its goal is to reveal underlying structures that escape close reading analysis and are supposed to be more important that the superficial structures visible in the traditional maps of linguistic atlases \cite
{goebl_regards_2003,goebl_sur_2008}.

The dialectometric work of Dees or Goebl have been mostly founded on the listing of lexical, phonological or morpho-syntactical traits (
``taxation'' \cite
{goebl_regards_2003}), and the analysis of the resulting data. 
The atlases produced by Dees' team \cite{dees_atlas_1980,dees_atlas_1987} so include a series of maps that each present a quantified opposition between two groups of forms, and can be used 
\cite
{goebl_sur_2008,dees_atlas_1980} as a matrix for computational analysis, 
both to study the underlying structures of dialectal variation or to locate a new text by confrontation with the already localised ones
or to cartography similarities between regions and map dialectal areas
\cite{goebl_lamenagement_2011,dees_atlas_1980,dees_dialectes_1985,dees_atlas_1987}.

The work of Dees and his Amsterdam School 
 and, after him, of Goebl and the Salzburg School, have given the rise to a 
more systematic and objective way to study medieval \textit{scriptae} (for an historical synthesis, 
see Volker \cite[chap.~2, p.~9-79]{volker_skripta_2003}).
Yet, an issue of circularity might still exist, since previous analyses usually based themselves on the localisation assigned to witnesses to identify linguistic areas and scriptological features. 
I would like to suggest a less supervised approach to the scriptometric analysis of the witnesses of a specific Old French epic genre, the \textit{chansons de geste}. My aim will be to identify main divisions in the corpus and to create profiles for each of them, and to verify both customary separations between \textit{scriptae} 
and the belonging of each individual witness to one of them.

\section{Corpus and Method}

In order to limit biases caused by stylistic, thematic or generic variations, this study will be limited to a single genre, the \textit{chansons de geste}. 
	Previous exploratory analyses, not shown here, 
	on a multi-generic corpus of 299 texts, did confirm that generic differences 
	interacted with linguistic boundaries and created too much noise.
Authorship related biases are hard to avoid, but might be counteracted by the very graphic variation observed in the witnesses, a problem in the stylometric analysis of medieval vernacular texts.  
The corpus of \textit{chansons} used here is composed of 50 witnesses (see app.~\ref{app:corpus}), with 1\,104\,296 tokens (geometric mean, 12\,016, median, 11\,490; min., 387; max., 217\,942).
The tokens are distributed between 52\,202 forms (long-tail distribution, with 25\,811 hapaxes; geom. mean of 2,57 occurrences, median, 2; 3\textsuperscript{rd} quartile, 4).
Editions were chosen for their use of a base witness (``copy-text'') --~the emphasis here being on the witnesses and not on the original text~-- as well as for their availability in digital form. The selection of witnesses was done empirically to have the largest corpus with a representativity of several putative regions of origin. Yet, its heterogeneity is a limitation\footnote{%
	I intend to work, in the coming years, on the constitution of a corpus as exhaustive as possible of epic witnesses (transcriptions, critical editions, manuscript descriptions). 
	The first few texts, encoded in \textsc{tei xml}, are available on Github \cite{Geste_2016}.
	The data, in \texttt{csv}, used for this paper, are available with scripts to reproduce analysis, on the Zenodo repository.
}.

Variation in editorial practice regarding the allographs \textbf{i}/\textbf{j} or \textbf{u}/\textbf{v} and their transcription led me to map all of them on \textbf{i} and \textbf{u}.
More generally, to avoid interferences with paleographic variation and perform on the graphematic level, all allographs (including ``capitals'') were normalized and all abbreviations expanded. The latter might be problematic, as it makes the process dependent on the choices of the editors, and can induce a bias, given that the norm is to use the majority unabbreviated form for expansion, inducing a distorsion favorable to this majority form as compared to the coexisting alternative ones \cite[p.~33]{morin_histoire_2007}.

It is to be noted that the exclusion of allographetic variation is an important simplification of the reality of textual witnesses, done both for contextual (the unavailability of consistent information) and theoretical reasons, based on the assumption that the variation in use of variant letter forms is more dependent on scribe's idiosyncrasies or script variation (\textit{textualis}, \textit{cursiva}, etc.), sometimes termed ``scribal mode'' \cite
{McIntosh_Scribal_1975,McIntosh_Towards_1974}. 
In the 
terminology offered by McIntosh for his ``scribal profiles'', 
this means we will restrict ourselves to the ``linguistic'' by opposition to the ``graphetic'' components \cite{McIntosh_Scribal_1975}, that is ``graphematic'', opposed to ``allographetic'' in the terminology retained here \cite{stutzmann_paleographie_2011}. 
Yet, given the interest of this latter kind of variation for dating and localising witnesses or identifying scribes, I have undertaken elsewhere to build a corpus of allographetic transcriptions and analyse them using similar techniques\footnote{%
	More details 
	can be found in 
	 \cite[chap.~2]{Camps_Otinel_2016}, including unsupervised clustering 
	 and allographetic profiles (sect.~2.4), 
	with a digital appendix 
	 giving access to the datasets and 
	 analysis procedures.
	An updated version of the corpus is available in 
	 \cite{Geste_2016}.
}.
Another dimension of these witnesses that we will not take into account concerns the alterations to the content of the text during its transmission (variants), that is the way in which the behaviour of the scribe alters the text of his model to result in a new copy, that we could term the ``diasystemic'' component, after Segre's definition \cite{segre_diasysteme_1976}.




If previous scriptometric works were based on the ``taxation'' of a defined list of features, I chose to use a bag-of-words approach on the graphic forms of the texts, in order to avoid inducing \textit{a priori} the features of the profiles. The main drawback 
 is that occurrences of an identical phenomenon (e.g. graphs of a given diphthong
) will be divided between all the forms that attest it. 
It will also prevent any syntactic feature to be taken into account and will limit the analysis to graphic or morphologic features. On the other hand, more limited habits, on the particular graph of a given lemma, will be fully accounted for
. Lexical variation, important for the localisation of texts through the identification of regional words \cite[p.~93]{duval_francais_2009}, will also be analysed this way, even if it makes the analysis highly dependent on content-based variation.
For this last reason, the database will be constituted of word rather than n-grams frequencies.


To limit content-based biases (and issues related to the non-Gaussian distribution of word-frequencies), only the most frequent words (MFW) are retained for analysis, an approach common in stylometry as well. Proper names were removed. 
This selection also leads to focusing the analysis on the dominant linguistic stratum (scribal or otherwise). Since no precise guidelines exist on the number of MFW to retain, robustness of the results will be checked with different levels of selection.

To cluster the witnesses
, hierarchical clustering was retained, a common analysis in scriptometrics \cite{goebl_regards_2003,goebl_lamenagement_2011}.
We do not yet possess guidelines on the effectiveness of various linkage criteria or distance measures in this field. Experimenting with a variety of those, to retain the one that would seem the best to me, though a heuristic approach advocated by Goebl \cite[p.~85]{goebl_regards_2003}, would induce a 
validation bias. As a consequence, I retained Ward's method, because it relies on the barycentre of the data clouds and allows for the constitution of balanced and coherent clusters,  often referred to as \textit{types}, as it minimises intra-cluster variation and maximises inter-cluster variation \cite{strauss_generalising_2017}. 
It is usually claimed that only squared euclidean distance is correct to use with Ward's linkage, because it relies on computations in euclidean space. Yet, recent research by Strauss and von Maltitz \cite{strauss_generalising_2017} seems to demonstrate that it can be generalised to use with Manhattan distance, and that this metrics outperforms euclidean in what regards the classification of (indo-european) languages, 
a statement that agrees 
with 
previous research in computational phonology applied to the clustering of (Dutch) dialects \cite{nerbonne_measuring_1997}, or with the supposed greater efficiency of Manhattan distance with highly dimensional data.

\section{Results}


\begin{figure} \centering
	\includegraphics[width=\textwidth]{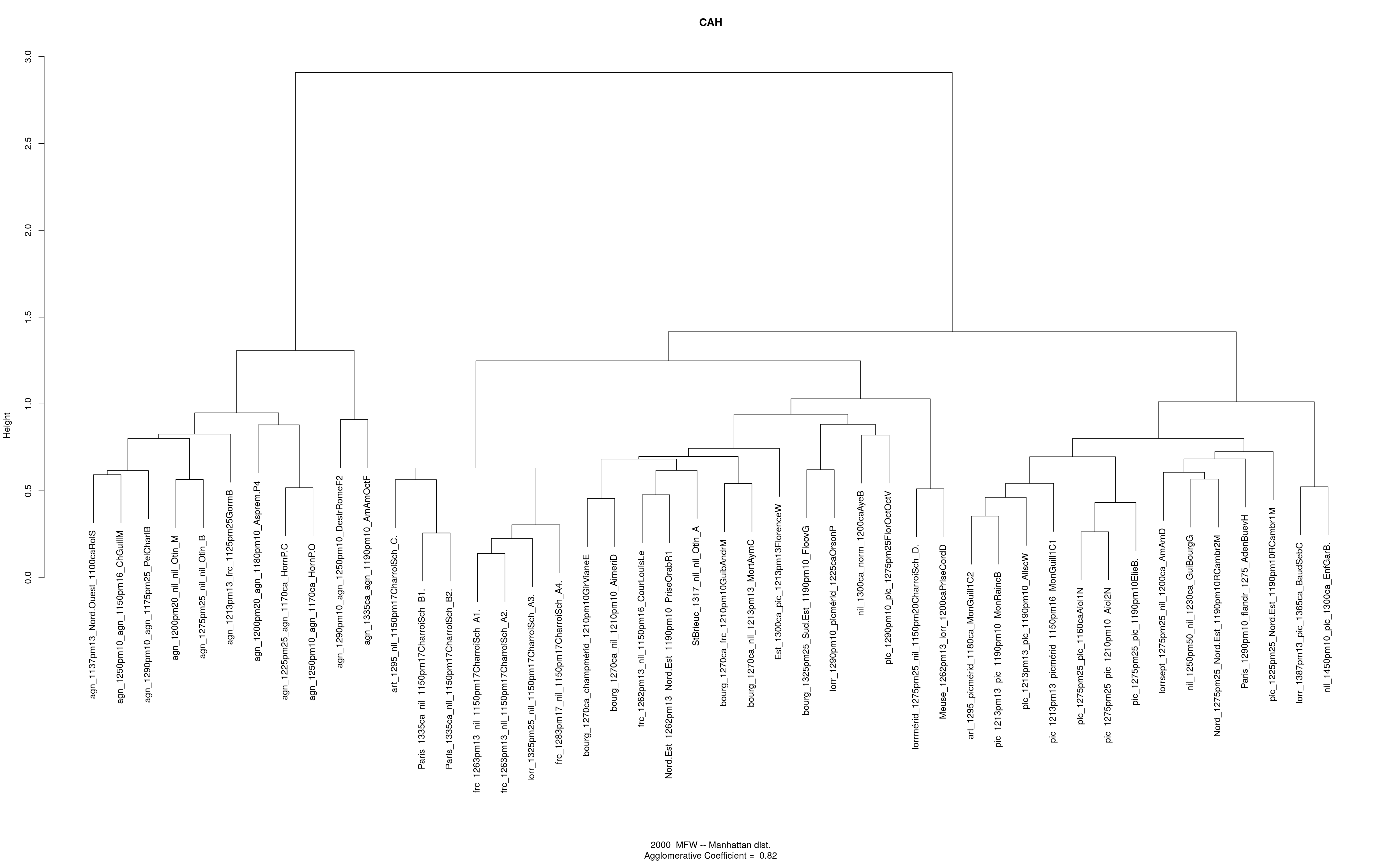}
	\caption{Hierarchical clustering of the \textit{Geste} corpus (Ward's method, Manhattan dist., 2000 MFW, relative freq.)}
	\label{fig:lang:geste:nca} %
\end{figure}

Results 
were mostly stable with between 600 and 3000 MFW, as well as the agglomerative coefficient (between $0.83$ and $0.8$). 
The main divisions (fig.~\ref{fig:lang:geste:nca}) are consistent with scriptological knowledge\footnote{%
	Following preliminary experiments, a few too short ($<$2000 words) witnesses were removed, because their inclusion tended to slightly twist the analysis. Nonetheless, their placement was consistent with the rest of the clustering: 
	Asprem\_C was placed in the Anglo-Norman cluster, among witnesses from the middle of the XIII\textsuperscript{th} century, at an intermediary position between witnesses of earlier or later texts, just on the left of MacaireAl2B, whose placement was also consistent with chronology; the CharroiSch\_fragm was in the Southern Lorraine group, with CharroiSch\_D and PriseCordD; Fier\_V was in the Lorraine/Burgundy group. See the online appendix.
}. The first opposes supposedly Anglo-Norman witnesses to Continental ones.  
Inside the Anglo-Norman group, a division opposes older (XII or XIII\textsuperscript{1/2}) to more recent (XIII-XIV) witnesses, arranged in an imperfect chronological order. The orientation is in itself interesting as it seems to confirm the hypothesis that later Anglo-Norman texts, written in a fossilising linguistic context, were more subject to continental norm. \enlargethispage{1\baselineskip}
The diachronic division of the Anglo-Norman group might also reveal the weakness of diatopic variation in this \textit{scripta}, in a country where ``\textit{Normannica lingua, que adventitia est, univoca maneat penes cunctos}'' (Ranulf Higden,  \textit{Polycronicon}, lib.~I, cap.~59). 
The second division, considerably lower, creates a separation within the continental groups, namely dividing Picard  witnesses of Picard texts from the rest.

The third division isolates mostly Central witnesses, but might also be due to authorial attraction between copies of the same text, that are even distributed between witnesses of the \textit{A}, \textit{B} and \textit{C} versions (not the \textit{D}). 
This might nonetheless have a linguistic sense, since \textit{A1} and \textit{A2} (and probably \textit{A4}), for instance, are known to come from the same workshop \cite[p.~434-436]{tyssens_typologie_1990}, as well as \textit{B1} and \textit{B2}.

Inside the group containing the rest of the Continental witnesses, which are mostly Eastern (or Lotharingian), divisions are weaker. Nonetheless, three subgroups can be individuated: witnesses from southern Lorraine (right), Burgundy (left), and Lorraine (centre).  Many of the apparent exceptions can be explained and concern witnesses whose origin is subject to debate or need rectification.
A subgroup of witnesses from Northern Lorraine or North-East appeared in the centre of this subgroup on some of the analyses (AmAmD, GuiBourG, RCambr), but are here grouped with Picard witnesses, maybe because one of them (RCambr) is a Northern copy of a text from the North-East. 

\begin{table}
	{\tiny %
		\hspace*{-2cm}%
		\begin{tabular}[t]{ccccccc}
			& v.test & mean in cat. & overall mean & sd in cat. & overall sd &  p.value\\ \hline \hline
			\multicolumn{7}{c}{Group 1 (\textit{Anglo-Norman})}\\ \hline
		pur&5.8438&0.0067&0.0018&0.0026&0.0032&0\\
		sunt&5.7222&0.0058&0.0016&0.0024&0.0028&0\\
		ad&5.6188&0.0120&0.0031&0.0056&0.0060&0\\
		mei&5.5343&0.0019&0.0005&0.0010&0.0010&0\\
		sur&5.5101&0.0044&0.0012&0.0021&0.0022&0\\
		lur&5.4663&0.0040&0.0010&0.0021&0.0021&0\\
		tut&5.4522&0.0045&0.0012&0.0023&0.0023&0\\
		al&5.3361&0.0072&0.0022&0.0034&0.0036&0\\
		e&5.3131&0.0357&0.0108&0.0127&0.0179&0\\
		sun&5.2683&0.0070&0.0018&0.0041&0.0037&0\\
		seit&5.2186&0.0020&0.0006&0.0012&0.0011&0\\
		dunt&5.1968&0.0018&0.0005&0.0011&0.0010&0\\
		od&5.1781&0.0033&0.0009&0.0019&0.0017&0\\
		si&5.1214&0.0186&0.0136&0.0030&0.0037&0\\
		mun&5.0508&0.0018&0.0005&0.0012&0.0010&0\\
		funt&5.0045&0.0008&0.0002&0.0006&0.0005&0\\
		reis&4.9249&0.0046&0.0012&0.0033&0.0026&0\\
		seignurs&4.9082&0.0009&0.0002&0.0006&0.0005&0\\
		rei&4.8912&0.0038&0.0010&0.0027&0.0022&0\\
		a&-4.8186&0.0246&0.0328&0.0050&0.0065&0\\
		droit&-4.8320&0.0001&0.0009&0.0002&0.0006&0\\
		qui&-4.8793&0.0037&0.0101&0.0032&0.0050&0\\
		mon&-4.9032&0.0003&0.0023&0.0006&0.0015&0\\
		et&-4.9212&0.0093&0.0352&0.0195&0.0201&0\\
		sont&-4.9557&0.0003&0.0028&0.0009&0.0019&0
		\end{tabular}
		\begin{tabular}[t]{ccccccc}
			& v.test & mean in cat. & overall mean & sd in cat. & overall sd &  p.value\\ \hline \hline
			\multicolumn{7}{c}{Group 4 (\textit{Picard})}\\ \hline
			ains&5.6322&0.0016&0.0005&0.0003&0.0007&0\\ 
			tous&5.4891&0.0021&0.0006&0.0006&0.0010&0\\ 
			passes&5.2743&0.0002&0.0000&0.0001&0.0001&0\\ 
			chou&5.2216&0.0009&0.0002&0.0006&0.0005&0\\ 
			trestous&5.0875&0.0003&0.0001&0.0001&0.0002&0\\ 
			tout&5.0120&0.0043&0.0015&0.0010&0.0020&0\\ 
			sarrasins&4.9654&0.0004&0.0001&0.0003&0.0002&0\\ 
			sains&4.9536&0.0004&0.0001&0.0002&0.0002&0\\ 
			toutes&4.9496&0.0004&0.0001&0.0001&0.0002&0\\ 
			commanda&4.9074&0.0001&0.0000&0.0001&0.0001&0\\ 
			cha&4.9023&0.0006&0.0001&0.0004&0.0003&0\\ 
			mieus&4.8405&0.0004&0.0001&0.0003&0.0002&0\\ 
			ochis&4.7118&0.0002&0.0000&0.0002&0.0001&0\\ 
			no&4.6579&0.0005&0.0002&0.0004&0.0003&0\\ 
			lieu&4.6264&0.0002&0.0001&0.0002&0.0001&0\\ 
			uausist&4.6239&0.0002&0.0000&0.0001&0.0001&0\\ 
			espiel&4.6180&0.0004&0.0001&0.0003&0.0002&0\\ 
			laissa&4.6063&0.0001&0.0000&0.0001&0.0001&0\\ 
			dolans&4.5675&0.0003&0.0001&0.0002&0.0002&0\\ 
			chi&4.5667&0.0009&0.0003&0.0006&0.0005&0\\ 
			toute&4.5588&0.0009&0.0004&0.0002&0.0004&0\\ 
			cief&4.4868&0.0007&0.0002&0.0005&0.0004&0\\ 
			ainc&4.4662&0.0008&0.0002&0.0005&0.0004&0\\ 
			mais&4.4656&0.0052&0.0023&0.0014&0.0023&0\\ 
			ceual&4.4543&0.0006&0.0002&0.0005&0.0003&0
		\end{tabular}
	}
	\caption{Scriptometric profiles for the Anglo-Norman (left) and Picard groups (right, without the Northern Lorraine subgroup), giving the 25 most characteristic forms (in positive or negative), rounded to 4 decimals}
	\label{lang:tab:profilsANetPic} %
\end{table}

Once groups are constituted, linguistic profiles for each of them can be built, at different levels, by estimating which features are the most characteristic with the values-test described by Lebart, Morineau et Piron \cite[p.~181-184]{Lebart_statistique_1995}%
\footnote{%
	The values-test is done by comparing $\bar{X_k}$, the mean
	of variable $X$ in category $k$ to the overall mean $\bar{X}$,
	while taking into account the variance $s_k(X)$ of this variable inside the class:
	$
	t_k(X) = \frac{
		\bar{X_k} - \bar{X}
	}{
		s_k(X)
	}
	$.
}, giving us an insight as to how clusters were constituted. To do so, the \texttt{catdes} function of the \texttt{FactoMineR} package by Francois Husson will be used.

The profiles for Anglo-Norman (table~\ref{lang:tab:profilsANetPic}) shows known features of this \textit{scripta}, like ``the replacement of Standard Medieval French (SMF) \textit{o} or \textit{ou} in all positions by \textit{u}'', ``the retention of \textit{ei} where \textit{SMF} develops \textit{oi}'', and ``the retention of dentals in 12\textsuperscript{th}-century texts''\cite[p.~45-46]{short_manual_2007}. Some are not usually cited: the use of \textit{e} (not \textit{et}), for instance, or \textit{al} (not \textit{au}).
The Picard group is also distinctively characterized by its palatalizations, its possessive of 1st and 2nd pers. pl. without -\textit{s} at the singular regime case or nominative plural  (\textit{no}, \textit{vo}), the use of \textit{tout}/\textit{tous} (not \textit{tuit}) at the masc. pl. nom., as well as the feminine \textit{toutes}, or the finales in -\textit{s} instead of -\textit{z}.

\section{Further research}

For the future of this research, an important aspect is the constitution of a corpus more homogeneous in terms of editorial practice. The extension of the corpus, by the addition of new witnesses, would make possible more focused analyses, with, for instance, more restricted chronological limits. 
The study of the relevance, both from a mathematical and philological point of view, of other metrics, is also a lead for future improvements.
It has been shown here, that, though interesting results on the grouping of the witnesses of literary texts can be obtained, their stratified nature remains an obstacle, causing some witnesses to switch groups according to either the presumed \textit{scripta} of their scribe, or the language of the author of the original text. Finding a more satisfying way to account for this phenomenon would be paramount to the scriptometric study of the tradition of medieval literary texts.

\appendix

\section{Corpus} \label{app:corpus}

\begin{footnotesize} \linespread{0.9}
	Sources: \textsc{and} $=$ \textit{Anglo-Norman Source Texts}, ed. David A. Trotter, William Rothwell, Geert De Wilde, and Heather Pagan, Aberystwyth and Swansea, 2001, \url{http://www.anglo-norman.net/sources/}.
	\textsc{Geste} \cite{Geste_2016}. \textsc{nca} $=$ \textit{Nouveau Corpus d’Amsterdam: corpus informatique de textes littéraires d’ancien français (ca 1150-1350)}, ed. Anthonij Dees, Achim Stein, Pierre Kunstmann, and Martin Dietrich Gleßgen, Stuttgart,  \url{http://www.uni-stuttgart.de/lingrom/stein/corpus}.
	\textsc{ota} $=$ \textit{The University of Oxford Text Archive}, ed. University of Oxford IT Services, s. d., \url{http://ota.ox.ac.uk/}.
	\textsc{tfa} $=$ \textit{Textes de français ancien}, ed. Pierre Kunstmann and Mark Olsen, 2003, Ottawa, \url{http://artfl-project.uchicago.edu/content/tfa}.
	\textsc{WikiS} $=$ \textit{Wikisource}, ed. Wikimedia Foundation, \url{http://en.wikisource.org/}.

	 We follow, when they exist, the identifier given in Möhren, Frankwalt, and Miller, Elena, 2010, \textit{DEAFBiblEl}, Heidelberg, \url{http://www.deaf-page.de/bibl_neu.php}..
\end{footnotesize}


\begin{tiny} \centering
\begin{tabular}{llllllll} 
\textbf{Source}&\textbf{DEAF}&\textbf{ms base}&\textbf{Ed}&\textbf{placeWit}&\textbf{dateWit}&\textbf{placeText}&\textbf{dateText}\\ \hline \hline  
TFA&AdenBuevH&Ars. 3142&Henry, 1953&Paris&1290pm10&flandr&1275\\  
OTA&AimeriD&BL Roy. 20 B.XIX&Demaison, 1852&bourg&1270ca&nil&1210pm10\\  
NCA+TFA&Aiol1NDeb&BnF fr. 25516&Normand et al., 1877&pic&1275pm25&pic&1160ca\\  
TFA&Aiol2N&BnF fr. 25516&Normand et al., 1877&pic&1275pm25&pic&1210pm10\\  
OTA&AliscW&Ars. 6562&Wienbeck et al., 1903&pic&1213pm13&pic&1190pm10\\  
NCA+TFA&AmAmD&BnF fr. 860&Dembowski, 1969&lorrsept&1275pm25&nil&1200ca\\  
AND&AmAmOctF&BL Roy. 12 C.XII&Fukui, 1990&agn&1335ca&agn&1190pm10\\  
GESTE&Asprem C&Clerm.-Fer. AD 1F2&Camps&agn&1250pm16&agn&1180pm10\\  
GESTE&Asprem P4&BnF, NAF 5094&Albarran \& Camps&agn&1200pm20&agn&1180pm10\\  
NCA&AyeB&BnF fr. 2170&Borg, 1967&nil&1300ca&norm&1200ca\\  
TFA&BaudSebC&BnF fr. 12552&Crist, 2002&lorr&1387pm13&pic&1365ca\\  
NCA&CharroiSch A1*&BnF fr. 774&Schoesler&frc&1263pm13&nil&1150pm17\\  
NCA&CharroiSch A2*&BnF fr. 1449&Schoesler&frc&1263pm13&nil&1150pm17\\  
NCA&CharroiSch A3*&BnF fr. 368&Schoesler&lorr&1325pm25&nil&1150pm17\\  
NCA&CharroiSch A4*&Trivulz. 1025&Schoesler&frc&1283pm17&nil&1150pm17\\  
NCA&CharroiSch B1*&BL Royal 20D XI&Schoesler&Paris&1335ca&nil&1150pm17\\  
NCA&CharroiSch B2*&BnF fr. 24369-70&Schoesler&Paris&1335ca&nil&1150pm17\\  
NCA&CharroiSch C*&Boul.-s.-M., BM 192&Schoesler&art&1295&nil&1150pm17\\  
NCA&CharroiSch D*&BnF fr. 1448&Schoesler&lorrmérid&1275pm25&nil&1150pm20\\  
NCA&CharroiSch fr.*&BnF NAF 934&Schoesler&nil&1250pm50&nil&1150pm17\\  
TFA&ChGuillM&BL Add. 38663&McMillan, 1949&agn&1250pm10&agn&1150pm16\\  
TFA&CourLouisLe&BnF fr. 1449&Lepage, 1978&frc&1262pm13&nil&1150pm16\\  
AND&DestrRomeF2&Hann. IV.578&Formisano, 1990&agn&1290pm10&agn&1250pm10\\  
NCA&ElieB*&BnF fr. 25516&P. Bloem&pic&1275pm25&pic&1190pm10\\  
TFA&EnfGarB*&BnF fr. 1460 &A. Kostka, 2002&nil&1450pm10&pic&1300ca\\  
GESTE&Fier-V&BAV Reg. lat. 1616&Camps&StBrieuc&1317&nil&1190ca\\  
GESTE&FloovG&Montp., F. Méd. 441&Guessard, 1858&bourg&1325pm25&Sud-Est&1190pm10\\  
NCA&FlorenceW&BnF NAF 4192&Wallenskoeld, 1907&Est&1300ca&pic&1213pm13\\  
NCA&FlorOctOctV&Bodl. Hatton 100&Vollmoeller, 1883&pic&1290pm10&pic&1275pm25\\  
NCA&GirVianeE&BL Roy. 20 B XIX&Van Emden, 1977&bourg&1270ca&champmérid&1210pm10\\  
NCA&GormB&Brux., BR port. II 181&Bayot, 1931&agn&1213pm13&frc&1125pm25\\  
NCA&GuibAndrM&BL Roy. 20 B XIX&Melander, 1922&bourg&1270ca&frc&1210pm10\\  
GESTE&GuiBourgG&Tours, BM 937&Guessard, 1858&nil&1250pm50&nil&1230ca\\  
AND&HornP-C&Cambr. Ff.VI.17&Pope, 1955&agn&1225pm25&agn&1170ca\\  
AND&HornP-O&Bodl. Douce 132&Pope, 1955&agn&1250pm10&agn&1170ca\\  
GESTE&MacaireAl2B&fragm. Loveday&Baker, 1915&agn&1250pm50&nil&1250pm50\\  
TFA&MonGuill1C1&Ars. 6562&Cloetta, 1906&pic&1213pm13&picmérid&1150pm16\\  
TFA&MonGuill1C2&Boul.-s.-M., BM 192&Cloetta, 1906&art&1295&picmérid&1180ca\\  
TFA&MonRaincB&Ars. 6562&Bertin, 1973&pic&1213pm13&pic&1190pm10\\  
WikiS&MortAymC&BL Roy. 20 B.XIX&Couraye, 1884&bourg&1270ca&nil&1213pm13\\  
NCA&OrsonP&BnF NAF 16600&Paris, 1899&lorr&1290pm10&picmérid&1225ca\\  
GESTE&OtinC A&Reg. lat. 1616&Camps&StBrieuc&1317&Nord-Est?&nil\\  
GESTE&OtinC B&Bodmer 168&Camps&agn&1275pm25&Nord-Est?&nil\\  
GESTE&OtinC M&BnF NAF 5094&Camps&agn&1200pm20&Nord-Est?&nil\\  
Divers&PelCharlB&BL Roy. 16 E.VIII&Bonafin, 1987&agn&1290pm10&agn&1175pm25\\  
NCA&PriseCordD&BnF fr. 1448&Densusianu, 1896&Meuse&1262pm13&lorr&1200ca\\  
TFA&PriseOrabR1&BnF fr. 774&Régnier, 1986&Nord-Est&1262pm13&Nord-Est&1190pm10\\  
NCA&RCambr1M&BnF fr. 2493&Meyer et al., 1882&pic&1225pm25&Nord-Est&1190pm10\\  
NCA&RCambr2M&BnF fr. 2493&Meyer et al., 1882&Nord&1275pm25&Nord-Est&1190pm10\\  
NCA&RolS&Bodl. Digby 23&Segre, 1971&agn&1137pm13&Nord-Ouest&1100ca
\end{tabular}
\end{tiny}

\end{document}